\documentclass[lettersize,journal]{IEEEtran}
\usepackage{amsmath,amsfonts}
\usepackage{amsmath}
\usepackage{algorithm}
\usepackage{booktabs}
\usepackage{algpseudocode}
\usepackage{stfloats}
\usepackage[numbers,sort&compress]{natbib}
\usepackage{array}
\usepackage{graphicx}
\usepackage[labelfont=scriptsize,textfont=scriptsize]{subfig}
\usepackage{textcomp}
\usepackage{stfloats}
\usepackage{amsthm}
\usepackage{url}
\usepackage{verbatim}
\usepackage{graphicx}
\newtheorem{definition}{Definition}
\begin{document}

\title{
	An Efficient Privacy-aware Split Learning Framework for Satellite Communications
}

\author{Jianfei~Sun$^1$, Cong Wu$^{1}$, Shahid Mumtaz$^{2}$, Junyi Tao$^{3}$, Mingsheng Cao$^{4}$, Mei Wang$^{5}$, and Valerio~Frascolla$^{6}$\\
	\emph{$^1$Nanyang Technological University, Singapore; $^2$Stony Brook University, USA;\\ 
	$^3$University of Electronic Science and Technology of China, China;\\ 
		$^4$Nottingham Trent University, UK; $^5$Shandong University, China; $^6$Intel Deutschland GmbH, Germany}\\
	\texttt{\scriptsize \{jianfei.sun, cong.wu\}@ntu.edu.sg, jutao@cs.stonybrook.edu, cms@uestc.edu.cn, wangmeiz@sdu.edu.cn, dr.shahid.mumtaz@ieee.org, valerio.frascolla@intel.com}
	}

\maketitle

\begin{abstract}
	In the rapidly evolving domain of satellite communications, integrating advanced machine learning techniques, particularly split learning, is crucial for enhancing data processing and model training efficiency across satellites, space stations, and ground stations. Traditional ML approaches often face significant challenges within satellite networks due to constraints such as limited bandwidth and computational resources. To address this gap, we propose a novel framework for more efficient SL in satellite communications. Our approach, Dynamic Topology-Informed Pruning, namely DTIP, combines differential privacy with graph and model pruning to optimize graph neural networks for distributed learning. DTIP strategically applies differential privacy to raw graph data and prunes GNNs, thereby optimizing both model size and communication load across network tiers. Extensive experiments across diverse datasets demonstrate DTIP's efficacy in enhancing privacy, accuracy, and computational efficiency. Specifically, on Amazon2M dataset, DTIP maintains an accuracy of 0.82 while achieving a 50\% reduction in floating-point operations per second. Similarly, on  ArXiv dataset, DTIP achieves an accuracy of 0.85 under comparable conditions. Our framework not only significantly improves the operational efficiency of satellite communications but also establishes a new benchmark in privacy-aware distributed learning, potentially revolutionizing data handling in space-based networks.
\end{abstract}


\section{Introduction}
6G wireless technology envisions integrated ground-air-space three-dimensional networks that cover satellites, aerial platforms, and terrestrial nodes as a key component for providing seamless global broadband coverage and computing services~\cite{yang20196g}. This challenge calls for a rethinking of integrated ground-air-space systems for communication and computing. Despite many results on LEO satellite and aerial platform communication, integrated networks across space-air-ground impose technical challenges, necessitating innovative approaches. Substantial recent research has focused on LEO and aerial platform-based broadband access, multilayer satellite network-based distributed computing, satellite-assisted sensing and integrated communication, IoT-over-satellite, and air IoT. Satellite networks play a pivotal role in global data communication, supporting critical functions such as geospatial analysis, real-time surveillance, and ubiquitous connectivity~\cite{qian2021multi,zhu2018dual, Koumaras2019eucnc}.
These networks generate extensive data that, when analyzed using advanced machine learning (ML) models like Graph Neural Networks (GNNs)~\cite{qu2024mobile,wang2024graph,moorthy2024survey}, can provide significant insights into network optimization, traffic management, and predictive maintenance, enhancing communication efficiency, identifying potential network bottlenecks or failures, and ensuring the effective allocation of resources across the satellite network. For instance, a privacy-aware system could be crucial in scenarios where sensitive geospatial data from satellites needs to be processed and analyzed without exposing raw data, such as in defense or disaster management applications. However, data processing and model training in these distributed networks often encounter bottlenecks due to limited bandwidth, high latency in space-to-earth communication, and computational limitations in edge satellites~\cite{verbraeken2020survey}. These challenges highlight the need for innovative approaches to optimize the performance and efficiency of satellite communication systems.

Distributed learning techniques, by reducing bandwidth usage, improving efficiency~\cite{DBLPSong_Lyu,lin2024adaptsfl,qu2024trimcaching,lin2024splitlora}, and leveraging GNNs, allow a much more effective data processing, addressing both computational and communication challenges. However, in satellite communications, where bandwidth is a scarce and costly resource, the distributed learning approach of transmitting extensive model updates poses significant challenges~\cite{mammen2021federated}, even potentially causing high bandwidth consumption and increased latency due to the volume of data transmitted between satellites and a central server. Conversely, split learning (SL) emerges as a more viable solution for satellite networks.
By dividing a neural network into segments, initial processing is conducted on the satellite, with complex computations offloaded to ground stations~\cite{kodheli2020satellite,fang2024automated,lin2024splitlora,lin2023pushing,lin2024split,lin2023fedsn,lin2024adaptsfl,lin2024efficient}.

This method substantially reduces the need to transmit raw data, conserving bandwidth, enhancing data privacy, and decreasing latency. The SL approach of transmitting only intermediate representations instead of complete model updates offers a more bandwidth-efficient alternative, making it a critical strategy for optimizing data processing in satellite-to-earth communications.

GNNs are essential in satellite networks because they can model complex dependencies and patterns in graph-structured data. This capability is crucial for optimizing network topologies, resource allocation, and predictive maintenance~\cite{wang2021grouting}. GNNs can capture and analyze these intricate relationships, making them invaluable for processing satellite data in applications such as advanced weather forecasting, where understanding atmospheric data interactions is vital.

Hence, implementing SL for GNNs becomes essential~\cite{chiang2019cluster,hu2020open,hamilton2017inductive}, as it offers a distributed learning framework that aligns with the operational constraints of satellite networks, ensuring efficient data processing while maintaining the integrity of the satellite system's complex data structure.
However, a significant challenge lies in efficiently adapting GNNs to the SL paradigm within the constraints of satellite networks. The core issue revolves around the trade-off between model complexity and the operational limitations of the satellite infrastructure. Optimizing this trade-off requires innovative approaches to model training and execution, ensuring that the GNNs can operate effectively over the limited computational resources and intermittent connectivity that characterize satellite networks.

Despite advancements in SL, existing approaches have not effectively addressed the need for a privacy-aware SL framework specifically designed for GNNs in satellite networks~\cite{poirot2019split,vepakomma2018split,gao2020end,koda2019one,DBLPSong_Lyu}. The importance of our study resides in the capability to provide effective means to bridge this gap in research.
Implementing SL for GNNs in satellite networks presents several technical challenges. One primary concern is maintaining robust model performance while reducing computational demands and model size, considering the resource constraints of satellite networks. Ensuring data privacy and security is another critical aspect, as sensitive graph data originating from satellites have to be protected.
Additionally, the GNN must be effectively partitioned between satellites and ground stations to optimize data processing and minimize data transmission. The dynamic and heterogeneous nature of satellite data further complicates this process. Addressing these challenges requires an innovative approach that integrates these elements into an SL learning framework.

The motivation behind our research is therefore driven by the need for a tailored and effective SL framework for GNNs in satellite networks. This framework aims to surmount these multifaceted challenges, offering a solution that enhances computational efficiency, ensures data privacy, and effectively handles the complex data structures typical of satellite communications.

\begin{figure}[!t]
	\centering
	\includegraphics[width=0.8\linewidth]{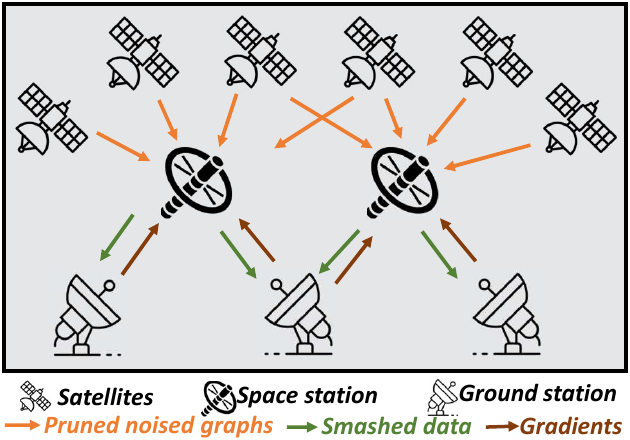}
	\caption{High-level system view of the satellite communication network and SL process. Satellites preprocess data and send smashed data to space stations, which perform intermediate computations and forward it to ground stations. Ground stations complete computations, update the global model, and send gradients and updates back, enabling iterative learning.}
	\label{fig:overview}
	\vspace{-6mm}
\end{figure}

\textbf{Our approach - DTIP.}
In this paper, we propose Dynamic Topology-Informed Pruning, namely DTIP, an efficient privacy-aware SL framework for GNN models in satellite networks.
DTIP leverages the synergy of differential privacy and strategic graph pruning to streamline both model complexity and the communication load between satellites and ground stations.
As illustrated in Figure~\ref{fig:overview}, DTIP prioritizes the structural importance of nodes and edges in a GNN while accommodating the variable computational capacities and connectivity challenges characteristic of satellite networks.
DTIP leverages the synergy of differential privacy and strategic graph pruning to optimize both the model size and the communication load between satellites and ground stations. The pruned and noise-processed graph data are then efficiently utilized in a SL setup.

In DTIP, SL between space stations and ground stations is executed with a unique approach. The space stations, equipped with a portion of the GNN model, perform initial computations on the pruned graph data, resulting in partial outcomes known as smashed data. This smashed data is then transmitted to the ground stations, which hold the remaining segments of the model. The ground stations complete the forward pass, compute the loss, and initiate the backpropagation process. The generated gradients are sent back to the space stations, enabling them to update their portion of the model. This collaborative process between space and ground stations ensures efficient model training while minimizing data transmission, which is essential in bandwidth-constrained satellite communication networks.

Note that we exclude direct satellite-to-ground transmission links in the system model depicted in Figure 1. This design choice is primarily due to bandwidth constraints, latency reduction, and computational offloading; space stations help manage bandwidth usage, reduce latency by aggregating data before transmission, and offload computational tasks from satellites. Additionally, space stations enhance network topology management, privacy and security measures, and scalability, enabling efficient routing, improved security, and streamlined communication paths as the number of satellites increases

\textbf{Novelty.} The novelty of DTIP lies in its unique integration of differential privacy and graph pruning within a novel SL framework, specifically designed for satellite communication networks. Our method adapts dynamically to fluctuating network topology and computational resources, a key feature for satellite environments. The synergy of differential privacy, graph pruning, and SL offers a distinct advantage by addressing the multifaceted challenges of satellite networks holistically. Specifically, (i) the dynamic topology-informed pruning mechanism adapts in real-time to changing network conditions, ensuring optimal performance, (ii) differential privacy adds controlled noise to raw graph data, safeguarding sensitive information, and (iii) graph pruning retains only the most critical nodes and edges, optimizing GNN performance in resource-constrained environments. SL enables distributed training, reducing data transmission needs, conserving bandwidth, and minimizing latency. These integrated techniques result in a more efficient, scalable solution that outperforms traditional ML methods by enhancing privacy, accuracy, and reducing computational and communication overhead, making DTIP highly effective for real-world satellite communication scenarios.

The contributions are be summarized as follows.
\begin{itemize}
	\item
	      We propose DTIP, the first privacy-aware split learning framework specifically designed for GNN models in satellite networks. By integrating differential privacy and graph pruning, DTIP optimizes model efficiency and data privacy. This framework uniquely addresses the computational and bandwidth challenges inherent in satellite communications.

	\item  DTIP innovatively integrates differential privacy, graph pruning, and GNN model pruning within a privacy-aware split learning context. This holistic method addresses critical challenges such as data privacy, computational efficiency, and limited bandwidth. DTIP achieves an optimal balance, preserving data utility, ensuring robust privacy, and reducing computational load.

	\item We extensively evaluate DTIP on four benchmark datasets, demonstrating its effectiveness in reducing computational burden and improving model performance under various conditions. Our results highlight DTIP's suitability for satellite network scenarios, consistently upholding data privacy and setting a new standard for privacy-aware distributed learning in resource-constrained environments.
\end{itemize}

\section{Related Work}
This section briefs related work on SL and model pruning.

\subsection{Split Learning}
FL is a distributed ML approach that allows training a global model across multiple clients, each with its local data, without sharing the data itself.
This technique involves clients performing local model training and then transmitting their local updates to a central server~\cite{mcmahan2017communication}.
The server aggregates these updates to form a global model and distributes the updated global parameters back to the clients for further training.
This process, often using algorithms like federated averaging (FedAvg), continues iteratively until the model converges~\cite{bonawitz2019towards,konevcny2016federated}.

SL, on the other hand, is a different paradigm for distributed learning where a neural network is divided into segments~\cite{vepakomma2018split,gupta2018distributed}.
The client-side network, operating on local data, computes activations up to a certain layer and sends these smashed data to the server.
The server completes the remaining computation, updates the network, and returns the gradients to the clients.
This split architecture allows the data to remain local, with only the necessary intermediate computations being communicated.
The learning can be synchronized across clients in either a centralized or peer-to-peer manner~\cite{gupta2018distributed,poirot2019split,vepakomma2018split}.

FL faces challenges, especially in large models where the amount of data exchanged between the clients and the server is proportional to the number of parameters in the network. This can make FL bandwidth-intensive and potentially impractical for models with a large number of parameters. To mitigate these issues, various improvements have been proposed to optimize the data exchanged during the learning process~\cite{sattler2019robust,wang2020optimize}.

\subsection{Model Pruning}

Model pruning optimizes neural networks, enhancing their speed and reducing their size with minimal loss in accuracy. In unstructured pruning, the network is refined by zeroing out less critical weights, resulting in a sparse weight matrix. This sparse representation maintains the efficacy of the network, even after considerable downsizing. However, this approach necessitates specialized computational resources that may not be accessible on basic devices~\cite{zhang2022advancing,han2015learning}.

Conversely, structured pruning takes a more holistic approach by excising entire convolutional filters, thereby narrowing the network's breadth. This method of pruning does not require unique computational capabilities, affording greater flexibility in deployment. Various criteria for selecting which filters to prune have been explored, including evaluating filters based on their L1 norms~\cite{li2016pruning} or the scaling factors within batch normalization layers~\cite{liu2017learning}. Innovations such as FPGM~\cite{he2019filter} and HRank~\cite{lin2020hrank} employ more nuanced criteria, aiming to enhance the pruning process.

The recent perspective on model pruning conceptualizes it as akin to network architecture search, intending to identify the most effective neural network configuration. Automated ML (AutoML) frameworks, such as AMC, utilize reinforcement learning to dynamically determine optimal pruning strategies~\cite{he2018amc}. Methods like MetaPruning~\cite{liu2019metapruning} and EagleEye~\cite{li2020eagleeye} adopt evolutionary algorithms and stochastic sampling to pinpoint the best possible network structure. These methods, while powerful, often demand significant computational investment due to the need for reiterative network training. ABCPruner~\cite{lin2020channel} addresses computational concerns by employing a more straightforward search method, yet it is not immune to the challenges of efficiency.

In GNNs, these pruning strategies must confront the unique limitations inherent to the complexity and irregularity of graphs data.
The adaptability and efficacy of pruning methods in GNNs remain areas requiring further exploration to overcome such challenges.

\section{Background}
\subsection{GNN for Satellite Communication}
GNNs, a subset of ML that has gained recently a lot of attention, are particularly adept at handling graph-structured data, which conventional neural networks often struggle with~\cite{hamilton2017inductive,hu2020open,wang2021grouting,chiang2019cluster}. Central to the functionality of GNNs is their ability to process and interpret the complex relationships inherent in graphs, making them ideal for analyzing network-based data. Therefore satellite networks, characterized by nodes (satellites, ground stations) and edges (communication links), can be effectively modeled and analyzed using GNNs. The underlying mechanism of GNNs involves a message-passing framework where the features of each node are iteratively updated based on the attributes of its neighboring nodes. Mathematically, this process can be expressed as \( h_v^{(l+1)} = \text{UPDATE}^{(l)} \left( h_v^{(l)}, \text{AGGREGATE}^{(l)} \left( \{ h_u^{(l)} : u \in \mathcal{N}(v) \} \right) \right) \), where \( h_v^{(l)} \) represents the feature of node \( v \) at the \( l \)-th layer, and \( \mathcal{N}(v) \) denotes its neighboring nodes.

In satellite communication systems, GNNs can be employed for various tasks, such as optimizing network routing, predicting link failures, and enhancing communication efficiency. Their ability to capture and analyze the spatial relationships between satellites enables more effective management of the network's dynamic topology. Furthermore, GNNs' applicability in node classification and link prediction tasks can help in proactive maintenance and efficient allocation of resources within the network. The advent of architectures like Graph Convolutional Networks (GCNs)~\cite{kipf2016semi} and Graph Attention Networks (GATs)~\cite{velivckovic2017graph} has further bolstered the use of GNNs in large-scale and complex networks.
Such architectures allow for scalable and adaptive learning, critical for addressing the challenges in the ever-evolving field of satellite communication networks.

\subsection{Split learning for  Satellite Network }

\begin{figure}[!t]
	\centering
	\includegraphics[width=0.75\linewidth]{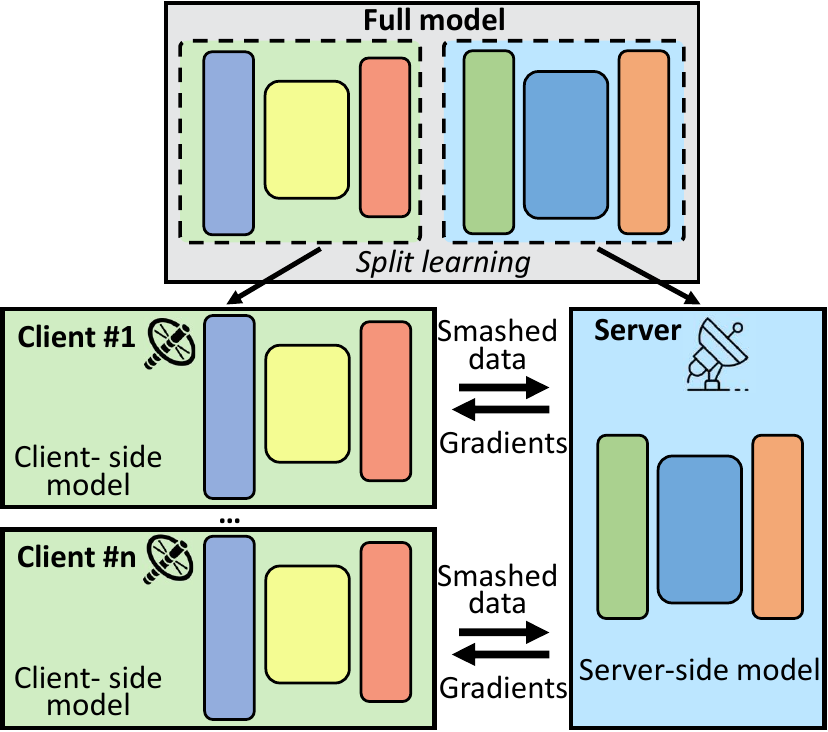}
	\caption{SL process detailed}
	\label{fig:split_learning_process}
\end{figure}

SL is emerging as one of the most innovative approaches in the domain of distributed learning, effectively partitioning a neural network into distinct segments distributed among various entities, typically comprising clients and a server~\cite{poirot2019split,vepakomma2018split,gao2020end}. SL facilitates collaborative training of deep neural networks, where clients, possessing private datasets, contribute to the learning process without the need to expose their raw data. As shown in Figure~\ref{fig:split_learning_process},
a client is responsible for the initial \( n \) layers \( f \) of the SL model, while the server manages the subsequent layers \( s \), culminating in a composite model \( F = s(f(\cdot)) \). The process entails clients sending the outputs from their initial layers \( f(X_{\text{priv}}) \) to the server, which then executes the forward pass and calculates the loss. Subsequent local optimization on \( s \) is performed by the server, which relays the gradient back to the client for back-propagation through \( f \). This technique of distributed back-propagation is particularly advantageous in scenarios demanding data privacy or operating under constrained computational resources, as it optimizes bandwidth usage while maintaining data confidentiality.

In satellite communications, SL finds a significant application, especially when integrated with GNNs. The satellite network, encompassing ground stations, space stations, and satellites, presents a unique challenge due to its dynamic and complex topology. By applying SL and GNNs, it is possible to efficiently manage and optimize this network. Ground stations can function as servers, handling complex computational tasks and coordinating the network's training. Space stations and satellites, acting as clients, manage portions of the neural network, processing data and contributing to the model's learning with reduced computational load. Such synergy between SL and GNNs enables enhanced data processing, predictive maintenance, and optimized resource allocation across the network. Additionally, the reduced bandwidth requirements of SL, compared to federated learning~\cite{qu2022blockchain,xu2023asynchronous}, make it highly suitable for satellite communications, where bandwidth is often a limiting factor. In fact, the ability to train models collaboratively without transmitting large volumes of data is crucial in maintaining efficient communication between earth-bound stations and satellites.

\subsection{Differential Privacy for Satellite Data}

Differential privacy ensures data privacy during statistical analysis, offering a mathematical safeguard against the identification of individuals' data~\cite{dwork2006differential}, particularly for sensitive satellite data. Satellites collect a wide array of information, ranging from environmental monitoring to geospatial intelligence.
Ensuring the privacy of such data is paramount, as it often includes sensitive or proprietary information. An
(\(\epsilon\), \(\delta\))-differential privacy approach to satellite datasets enables the safeguarding of individual data points within these vast datasets, while still allowing for meaningful statistical analysis.
It is defined with two parameters, \(\epsilon\) (privacy loss) and \(\delta\) (failure probability), which quantify the strength of the privacy guarantee.

\begin{definition}
	\label{def:dp}
	A mechanism \( M \) is (\(\epsilon\), \(\delta\))-differentially private if, for any pair of adjacent datasets \( x \) and \( y \), and for all possible outputs \( R \), the following inequality holds:
	\begin{align}
		P[M(x) \in R] \leq e^\epsilon \cdot P[M(y) \in R] + \delta.
	\end{align}
\end{definition}

Selecting \(\epsilon\) and \(\delta\) involves a trade-off: lower values mean higher privacy at the potential cost of data utility due to increased noise. The choice of these parameters should align with privacy risks and the expected level of data protection.
A smaller $\epsilon$ means better privacy, as it ensures that the presence or absence of a single individual in the dataset does not significantly affect the output. However, a very small $\epsilon$ can make the data less useful, as it requires adding more noise to the data to preserve privacy. Typical values for
$\epsilon$ can range from 0.01 for very strong privacy to around 1 for weaker privacy.
Paremeter $\delta$ represents the probability of privacy leakage.
Smaller $\delta$ means a lower chance of leakage, but again, too small a value can reduce the utility of the data. Common practice is to set $\delta$ to a value less than the inverse of the population size in the dataset.
For example, for a dataset with $1,000,000$ entries, $\delta$ might be set to less than $\frac{1}{1,000,000}$.

\section{DTIP Framework Design}

In this section we describe our proposed system architecture and provide details on its components and related technologies.

\begin{figure}[!t]
	\centering
	\includegraphics[width=\linewidth]{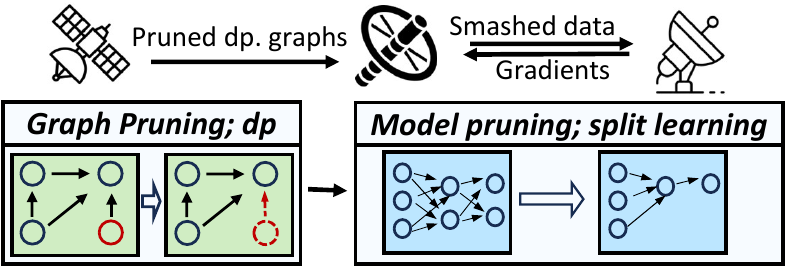}
	\caption{Overview of the system design}
	\label{fig:system_overview}
\end{figure}
\subsection{System Overview}
Our framework optimizes model accuracy, data privacy, and computational efficiency by balancing these three objectives. This involves trade-offs between high performance, robust privacy, and reduced computational load. Our Dynamic Topology-Informed Pruning (DTIP) framework adapts dynamically to network conditions, ensuring efficient and effective performance in satellite communication networks.

The design of our system centers on the integration of differential privacy and graph pruning methodologies within a SL framework, tailored specifically for GNNs, as shown in Figure~\ref{fig:system_overview}. This innovative approach addresses the challenges of data privacy, computational efficiency, and network integrity in distributed learning environments, particularly in scenarios involving complex graph-structured data. By harmonizing advanced privacy-preserving techniques with strategic model optimization, our system sets a new benchmark in distributed ML, especially in contexts where data sensitivity and processing capabilities are critical factors.

In the first facet of our system, differential privacy is
applied to graph-structured data, involving the careful calibration of noise addition to query outputs on graphs, ensuring robust privacy guarantees while retaining aggregate information utility. Complementing this privacy aspect is our innovative approach to graph and model pruning within the SL architecture. This dual pruning strategy, encompassing both graph sparsification and network weight reduction, is pivotal in managing the computational complexity of GNNs. Graph pruning reduces the intricacy of graph structures by targeting specific edges based on their contribution to network performance, while model pruning streamlines the neural network by selectively trimming less significant weights. Together, these methods enhance the efficiency of GNNs in distributed learning scenarios, optimizing them for environments where computational resources are at a premium, and ensuring a system that is both robust and agile.

\subsection{Differential Privacy for Graphs}
In privacy-preserving data analysis, differential privacy stands out as a robust framework. Its application to graph-structured data, which inherently features intricate relationships and interconnectivity, requires careful consideration. The core idea hinges on the adjacency of graphs: two graphs \( G_1 \) and \( G_2 \) are considered adjacent, denoted \( G_1 \sim G_2 \), if they differ by only one node or edge. This adjacency is crucial for defining the sensitivity of graph queries, mirroring the concept of neighboring datasets in traditional data types.

Sensitivity in graph contexts measures the maximum potential change in a query's output due to the modification of a single edge or node. Formally, the sensitivity \( \Delta f \) of a function \( f \) operating on a graph is defined as:
\begin{align}
	\Delta f = \max_{G_1 \sim G_2} | f(G_1) - f(G_2) |
\end{align}.
This metric is instrumental in determining the appropriate level of noise addition for achieving differential privacy.

\emph{Differential Privacy with Noise Addition:}
The Gaussian mechanism is often utilized for achieving \((\epsilon, \delta)\)-differential privacy. The standard deviation \( \lambda \) of the Gaussian noise is calculated as:
\begin{align}
	\lambda \geq \Delta f \cdot \sqrt{2 \ln(1.25/\delta)} / \epsilon,
\end{align}
ensuring that extreme deviations from the true value are bounded by \( \delta \), thus providing a strong privacy guarantee.

\emph{Output Perturbation:}
To attain a differentially private output \( f'(G) \) for a query function \( f \), Gaussian noise is added to the query result:
\begin{align}	f'(G) = f(G) + \mathcal{N}(0, \lambda) .
\end{align}
This noise addition ensures that any single alteration in the graph minimally impacts the output, safeguarding individual privacy.

\begin{algorithm}[!t]
	\scriptsize
	\caption{Apply \((\epsilon, \delta)\)-Differential Privacy to Graphs}
	\label{alg:applydp}
	\begin{algorithmic}[1]
		\Statex \textbf{Input:} Graph \( G = (V, E) \), Query function \( f \)
		\Statex \textbf{Output:} Differentially private approximation \( f'(G) \)
		\State \textbf{Parameters:} Privacy levels \( \epsilon > 0 \), \( \delta > 0 \)
		\State \textbf{define} \( \text{Adjacency}(G_1, G_2) \) \Comment{Determine adjacency}
		\State \( \Delta f \gets \max_{G_1 \sim G_2} | f(G_1) - f(G_2) | \) \Comment{Calculate sensitivity}
		\State \( \lambda \gets \Delta f \cdot \sqrt{2 \ln(1.25/\delta)} / \epsilon \) \Comment{Compute noise scale}
		\State \( \text{Noise} \gets \mathcal{N}(0, \lambda) \) \Comment{Generate Gaussian noise}
		\State \( f'(G) \gets f(G) + \text{Noise} \) \Comment{Add noise to query output}
		\State \textbf{return} \( f'(G) \) \Comment{Release differentially private result}
	\end{algorithmic}
\end{algorithm}
\emph{Probability Guarantees:}
From Definition~\ref{def:dp}, the \((\epsilon, \delta)\)-differential privacy model guarantees that for any two adjacent graphs \( G_1 \) and \( G_2 \), and any output set \( S \), the probability of the differentially private output falling in \( S \) is constrained, within a multiplicative factor of \( e^\epsilon \) and an additive term of \( \delta \):
\begin{align}
	P[f'(G_1) \in S] \leq e^\epsilon \cdot P[f'(G_2) \in S] + \delta.
\end{align}

Algorithm~\ref{alg:applydp} explains how to implement \((\epsilon, \delta)\)-differential privacy on graph-structured data, which ensures the privacy of individuals represented in the graph while allowing for the extraction of valuable aggregate information.
It commences by defining the adjacency criterion in the graph, which is essential to understanding how a single change (addition or removal of a node or edge) can impact the data. Sensitivity of the query is then computed, indicating the maximum effect a single adjacency change can have on the query output. Privacy levels are set using parameters \( \epsilon \) and \( \delta \), where \( \epsilon \) controls the strictness of the privacy guarantee and \( \delta \) allows a small probability of exceeding this guarantee. The algorithm calculates the scale of Gaussian noise based on the sensitivity and privacy parameters, ensuring that the added noise is proportional to the sensitivity and inversely proportional to the desired privacy level. This noise is then added to the query output, resulting in a differentially private release. The output, thus perturbed, maintains statistical indistinguishability, regardless of the presence or absence of any single individual in the dataset, achieving a balance between preserving individual privacy and retaining the utility of the data for analysis.

\subsection{Graph Pruning}

\emph{Node Importance Assessment.}
We utilize eigenvector centrality (EC) and betweenness centrality (BC) as metrics for assessing the importance of nodes in a satellite network graph.
	{EC} measures the influence of a node based on the centrality of its connections and is defined as:
\begin{align}
	\label{eq:ec}
	EC_v=\alpha \sum_{u}^{|{V}|}a_{v,u}EC_u,
\end{align}
where $\alpha$ is a positive constant.
$a_{v,u}$ is the element of the adjacency matrix that indicates
if node $v$ is connected to $u$.
$a_{v,u} = 1$ if node $v$ is linked to k,
and $a_{v,u} = 0$ otherwise.
	{BS} quantifies how often a node appears on the shortest paths between two other nodes and is calculated as:
\begin{align}
	\label{eq:bc}
	BC(v)=\frac{2}{(|V|-1)(|V|-2)}\sum_{i\neq j \neq v \in V} \frac{|  \operatorname{sp}(i,j|v)|}{ |  \operatorname{sp}(i, j)|},
\end{align}
where $|\operatorname{sp}(i, j)|$ is the total number of the shortest paths from node $i$ to $j$,
$\operatorname{sp}(i,j|v)$ is the number of those paths that pass through node $v$.
Nodes with high BC are crucial for network communication efficiency.
To compute the centrality
, we use the harmonic mean of EC and BC.

	{Harmonic mean} calculates the harmonic mean of EC and BC for each node in the graph. The harmonic mean is given as
$	H_{\text{mean}}(v) = \frac{EC(v) + BC(v)}{EC(v) *BC(v)} $, where \(EC(v)\) and \(BC(v)\) are the EC and BC of node \(v\), respectively.

\emph{Pruning Threshold Setting.}
To prune the input graphs using the measured harmonic mean of EC and BC we drop the nodes with low harmonic value using a pre-defined ratio in the graph.
In setting the pruning threshold, we determine it based on the distribution of the harmonic mean scores of the nodes.
Specifically, we set our threshold as a specific value defined using the mean ($\mu$) and standard deviation ($\sigma$) 
of harmonic scores.
The nodes that have an harmonic mean lower than such threshold
are candidates for pruning.
This approach ensures to retain the most crucial nodes for the network integrity and functionality, while removing those that contribute less to the overall structure and efficiency of the graph.

\subsection{Model Pruning}

GNNs have become increasingly important in processing and learning from graph-structured data. However, as GNN models grow in size and complexity, they also become computationally expensive, both in terms of memory and processing power. This is especially problematic in distributed learning environments, where computational resources and bandwidth may be limited. Model pruning in GNNs addresses these challenges by removing redundant or less significant parts of the model, thereby reducing its complexity without significantly compromising performance.

SL, as a specialized form of distributed learning, divides the neural network training across various nodes or devices. This setup is especially useful for ensuring data privacy and handling large datasets that are impractical to process centrally.
The pruning process in SL for GNNs involves a detailed mathematical approach aimed at optimizing the model's efficiency while retaining its accuracy. The process consists of two primary steps: graph sparsification and network weight pruning.

Graph sparsification involves reducing the complexity of the graph structure of the GNN. Mathematically, this is achieved by creating a pruning mask \( m_g \) for the adjacency matrix \( A \) of the graph \( G \). The mask \( m_g \) is designed to identify and remove edges that are less critical for the network performance. These include negative edges, which connect nodes with differing labels (denoted as \( E_{\text{neg}} \)), and non-bridge edges, which do not affect the connectivity of the graph when removed (denoted as \( E_{\text{non}} \)).

\begin{algorithm}[!t]
	\caption{GNN Pruning for Split Learning}
	\scriptsize
	\label{alg:gnnpruning}
	\begin{algorithmic}[1]
		\Statex \textbf{Input:} Graph $G = \{X, A\}$, Training labels $Y_t = \{y_1, y_2, \dots, y_t\}$, Desired pruning ratios $p_g$, Network mask $m_\Theta$, Split nodes $\{N_1, N_2, \dots, N_k\}$
		\Statex \textbf{Output:} Sparsified Graph mask $m_g$, Updated network weights

		\State Distribute $G$ and $Y_t$ across split nodes $\{N_1, N_2, \dots, N_k\}$
		\State Initialize local models on each node with shared global parameters
		\For{each split node $N_i$}
		\State Train local model $f_{m_{\Theta_i}} \odot \Theta_i (X_{N_i}, m_g \odot A_{N_i})$ with $Y_{t_{N_i}}$
		\State Compute local gradients and update global parameters
		\EndFor
		\State Aggregate predictions $Y^\prime$ from all split nodes
		\State Compute $A_{\text{neg}}(i, j) = m_g \odot 1[Y^\prime(i) \neq Y^\prime(j)]$ $\forall i, j \in \{1, \dots, |V|\}$
		\State Remove $p_g \times \|A\|_0$ of $A_{\text{neg}}$ by setting corresponding $m_g = 0$
		\If {$\|A_{\text{neg}}\|_0 == 0$}
		\State Find all non-bridges $A_{\text{non}}$
		\State Remove $p_g \times \|A\|_0$ of $A_{\text{non}}$ by setting corresponding $m_g = 0$
		\EndIf
		\State Communicate updated $m_g$ to all split nodes
		\For{each split node $N_i$}
		\State Update local model with new $m_g$
		\State Retrain and update global parameters if necessary
		\EndFor
	\end{algorithmic}
\end{algorithm}
\subsection{Split Learning for GNN}

Network weight pruning aims to reduce the complexity of the neural network model itself. It involves applying a network mask \( m_\Theta \) to the network weights \( \Theta \). The mask \( m_\Theta \) is computed based on the significance of each weight, often determined by metrics such as magnitude or gradient. The goal is to retain weights that are crucial for the model performance while pruning those that have minimal impact.

The combination of these two methods results in a pruned GNN model that is computationally efficient and suitable for distributed environments, such as SL setups, without significantly compromising the model's predictive accuracy.

Algorithm~\ref{alg:gnnpruning} shows the detailed workflow to function efficiently in distributed learning environments. Initially, the graph \( G = \{X, A\} \) along with the training labels \( Y_t \) are distributed across the split nodes, where each node initializes its local model with shared global parameters. These local models are then trained independently on each node. Post-training, the local gradients are computed and used to update the global model parameters. Following this, predictions from all nodes are aggregated to compute the pruning masks for both the graph and the network weights. The key focus during pruning is the removal of negative edges and non-bridges from the graph, leading to an updated graph mask \( m_g \). This updated mask is then communicated back to all split nodes, and each node updates its local model with this new graph structure. If necessary, the local models undergo a retraining phase with the updated graph to fine-tune the global model parameters.
The satellite-based distributed learning scenario involves a three-tiered architecture comprising satellites, space stations, and ground stations, each with varying computational capabilities, as shown in Figure 1 and detailed in Algorithm~\ref{alg:c1},~\ref{alg:c2}, and~\ref{alg:s}.

\emph{Satellites (Client-Side 1):} Satellites, equipped with basic computational resources, are responsible for initial data collection and processing. The key operation at this level is applying differential privacy	and graph pruning, denoted as $\text{PruneGraph}(G_{\text{satellite}})$, where $G_{\text{satellite}}$ represents the raw graph data. This process simplifies the graph structure, yielding a pruned graph $G_{\text{pruned}}$. The pruned graph data are then transmitted to the space stations for further processing.
\begin{algorithm}[!t]
	\caption{Satellites (Client-Side 1)}
	\label{alg:c1}
	\scriptsize
	\begin{algorithmic}[1]
		\Statex \textbf{Input:} $G_{\text{satellite}}$
		\Statex \textbf{Output:} $G_{\text{dp, pruned}}$

		\State $\Phi_{\text{satellite}}^{(G)} \leftarrow \text{InitializeSatelliteModel}()$
		\For{each graph data $G_{\text{satellite}}$}
		\State $G_{\text{dp}} \leftarrow \text{AddDifferentialPrivacy}(G_{\text{satellite}})$
		\State $G_{\text{dp, pruned}} \leftarrow \text{PruneGraph}(G_{\text{noisy}})$
		\State Send $G_{\text{noisy, pruned}}$ to space stations
		\EndFor
	\end{algorithmic}
\end{algorithm}

\emph{Space stations (Client-Side 2):} Space stations, possessing medium-level computational resources, receive the pruned graph data $G_{\text{pruned}}$ and perform the initial stages of GNN message propagation. For each node $v_i$ in $G_{\text{pruned}}$, the algorithm aggregates features from its neighbors $\mathcal{N}(v_i)$ using an AGGREGATE function, updating each node's representation to $h_{v_i}^{(l+1)}$. The updated node features for the entire graph batch, $H_{G_{\text{pruned}}}^{(l+1)}$, are then sent to the ground stations for advanced processing and model training.

\begin{algorithm}[!t]
	\scriptsize
	\caption{Space Stations (Client-Side 2)}
	\label{alg:c2}
	\begin{algorithmic}[1]
		\Statex \textbf{Input:} $G_{\text{dp, pruned}}$: Pruned graph data received from satellites
		\Statex \textbf{Output:} $H_{G_{\text{}}}^{(l+1)}$: Updated node features for the pruned graph

		\State $\Phi_{\text{space}}^{(G)} \leftarrow \text{InitializeSpaceModel}()$
		\Loop
		\State Receive $G_{\text{dp, pruned}}$ from satellites
		\State Receive $G_{\text{dp, pruned}}$ from ground stations
		\For{each node $v_i$ in $G_{\text{dp, pruned}}$}
		\State $h_{v_i}^{(l+1)} \leftarrow \text{AGGREGATE}^{(l)}\left(\left\{ h_{u}^{(l)} : u \in \mathcal{N}(v_i) \right\}\right)$
		\EndFor
		\State $H_{G_{\text{}}}^{(l+1)} \leftarrow \left\{ h_{v_i}^{(l+1)} : v_i \in G_{\text{dp, pruned}} \right\}$
		\State Send $H_{G_{\text{}}}^{(l+1)}$ to ground stations
		\EndLoop
	\end{algorithmic}
\end{algorithm}

\emph{Ground stations (Server-Side):} Ground stations, with the highest computational power, handle the final and most demanding phase of the learning process. They receive the processed node features $H_{G_{\text{pruned}}}^{(l+1)}$ and execute advanced model training and gradient optimization. The gradients, $\nabla \Phi_{\text{ground}}^{(G)}$, are computed and used to update the model $\Phi_{\text{ground}}^{(G)}$. The refined model or insights are subsequently disseminated back to the space stations and satellites, completing the feedback loop and enabling continual adaptation and improvement of the learning model.

\begin{algorithm}[!t]
	\caption{Ground Stations (Server-Side)}
	\label{alg:s}
	\scriptsize
	\begin{algorithmic}[1]
		\Statex \textbf{Input:} $H_{G_{\text{pruned}}}^{(l+1)}$: Processed node features from space stations
		\Statex \textbf{Output:} Updated model $\Phi_{\text{ground}}^{(G)}$ and model insights sent back to space stations and satellites
		\State $\Phi_{\text{ground}}^{(G)} \leftarrow \text{InitializeGroundModel}()$
		\Loop
		\State Receive $H_{G_{\text{pruned}}}^{(l+1)}$ from space stations
		\State $\Phi_{\text{ground}}^{(G)}.\text{train}(H_{G_{\text{pruned}}}^{(l+1)})$
		\State $\nabla \Phi_{\text{ground}}^{(G)} \leftarrow \text{Compute Gradient}(\Phi_{\text{ground}}^{(G)})$
		\State Update $\Phi_{\text{ground}}^{(G)}$ using $\nabla \Phi_{\text{ground}}^{(G)}$
		\State Send model updates back to space stations and satellites
		\EndLoop
	\end{algorithmic}
\end{algorithm}

\section{Performance Evaluation}
During evaluation, we aim to answer the following research questions:     \textbf{RQ1} How do different levels of differential privacy impact the overall performance of the system?
\textbf{RQ2} What is the effect of the proposed graph pruning technique on the GNN models accuracy and robustness?
\textbf{RQ3} How does the proposed model pruning approach impact the model performance?
\textbf{RQ4} How does the DTIP framework impact the computational efficiency of GNNs?

\subsection{Experimental Setup}

\emph{Dataset.} For our experimental evaluation, we utilized the following four real datasets: Amazon2M~\cite{chiang2019cluster},
Reddit~\cite{hamilton2017inductive},
Facebook~\cite{traud2012social},
ArXiv~\cite{hu2020open}.
Each dataset offers unique insights and challenges, aligning well with the diverse requirements of satellite-based distributed learning environments.

\begin{itemize}
	\item \emph{Amazon2M} dataset  is a vast Amazon co-purchasing network where each node represents a product with features derived from a bag-of-words model of the product description. The edges signify co-purchasing relationships between products. For our study, we partitioned it into a private graph of 1 million nodes and a public graph with the remainder.

	\item \emph{Reddit} dataset  captures post-to-post interactions, with edges indicating a shared commenter. We sampled 300 nodes per class for the private graph and selected 15,000 nodes each for public training and testing.

	\item \emph{Facebook} dataset is an anonymized network of Facebook users from various US universities. We focused on the network among University of Illinois Urbana-Champaign (UIUC) students, predicting their class year. The nodes and edges represent users and friendships, respectively, divided equally into private and public graphs.

	\item \emph{ArXiv} dataset  is a citation network where nodes are papers and edges citations. We constructed the private graph from papers published until 2017 and used papers from subsequent years for public training and testing.

\end{itemize}

Table~\ref{tab:Data} presents data statistics of the experimental datasets, where deg is the average degree of the graph.
We select 50\% of Amazon2M and Reddit, 20\% of Facebook, and  40\% of ArXiv dataset as the test set from the public graph.
In our experiments, we train the student model using a small subset of pseudo-labeled query nodes, limited to a maximum of 1000, regardless of the larger size of the public dataset.
\begin{table}[!t]
	\centering
	\caption{Data statistics of experimental datasets}
	\label{tab:Data}
	\resizebox{\linewidth}{!}{\begin{tabular}{p{30pt}p{23pt}p{20pt}p{25pt}p{25pt}p{20pt}p{25pt}p{25pt}p{20pt}}
			\hline 	\hline
			         &          &     & \multicolumn{3}{c}{Private} & \multicolumn{3}{c}{Public}                                     \\
			\cmidrule[0.8pt](rl){4-6}   \cmidrule[0.8pt](rl){7-9}
			         & \# class & $X$ & \# node                     & \# edge                    & deg   & \# node & \# edge & deg   \\
			\hline 	\hline
			Amazon2M & 47       & 100 & 1M                          & 12.7M                      & 12.73 & 1.4M    & 21.8M   & 15.08 \\
			Reddit   & 41       & 602 & 12300                       & 266148                     & 21.64 & 30000   & 876846  & 29.23 \\
			Facebook & 34       & 501 & 13401                       & 270992                     & 20.22 & 13402   & 266141  & 24.82 \\
			ArXiv    & 40       & 128 & 90941                       & 187419                     & 2.06  & 78402   & 107900  & 1.38  \\
			\hline
		\end{tabular}}
\end{table}

\emph{Implementations.}
In our experimental design, the core of the private and public models is constructed using a two-layer GraphSAGE architecture~\cite{hamilton2017inductive}, chosen for its effectiveness in learning on graph-structured data.
We use torch geometric to simulate SL on a devices equipped with RTX 3090 GPU.
GraphSAGE is an inductive learning framework that generates node embeddings by sampling and aggregating features from a node's local neighborhood. It updates node representations by aggregating features of neighbors, allowing for the generation of embeddings for unseen nodes. This approach is key for dynamic graphs like satellite networks, where adaptability to new nodes is essential.
Each layer has a hidden dimension of 128 and utilizes the ReLU activation function to introduce non-linearity. To facilitate model generalization and mitigate overfitting, we employed batch normalization on the output of the first layer and set a dropout rate of 0.3. The learning rate was selected at 0.01, so to optimize the trade-off between convergence speed and stability. For the optimization process, the Adam optimizer~\cite{kingma2014adam} 
was employed for its efficiency in handling sparse gradients and adaptive learning rate adjustments.

\subsection{Effectiveness of Applying Differential Privacy to Graphs}
\begin{figure}[!t]
	\centering
	\includegraphics[width=0.65\linewidth]{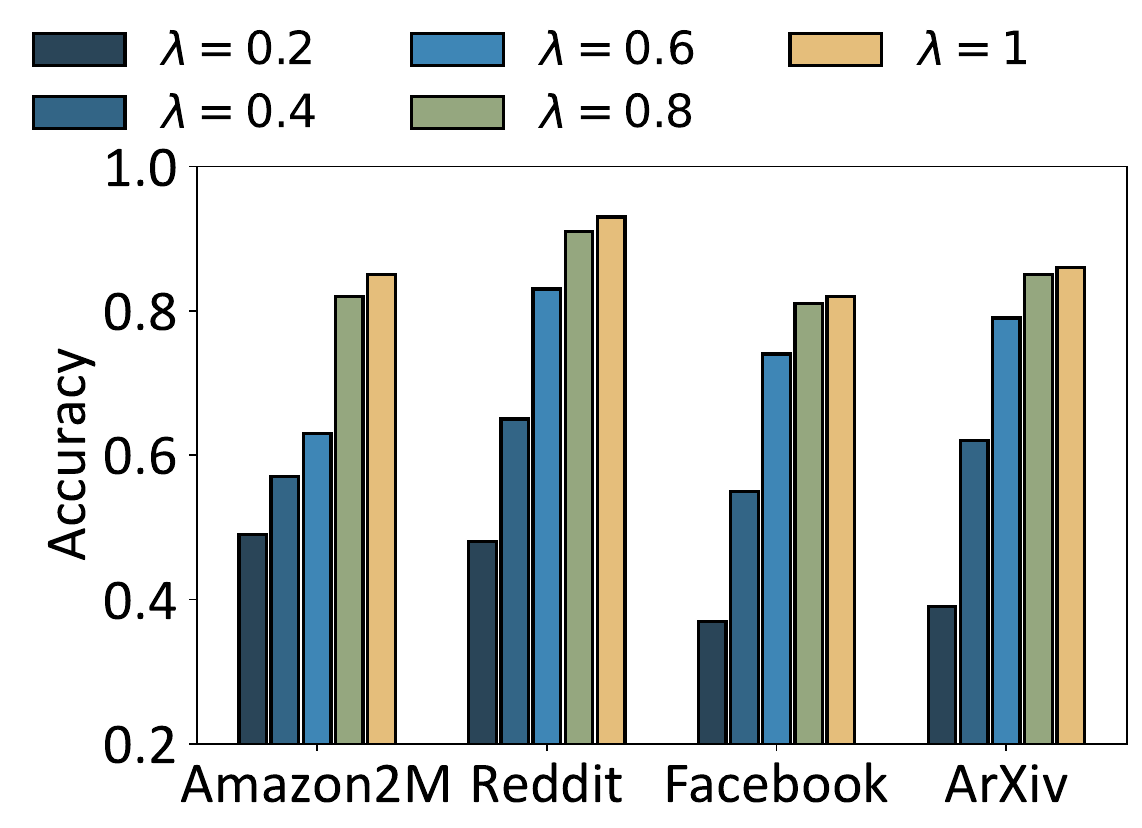}
	\caption{Accuracy vs. noise ($\propto \frac{1}{\epsilon}$) for different datasets}
	\label{fig:accuracy_dp}
\end{figure}

In assessing the privacy-utility trade-off, we varied the parameter $\lambda$ to observe its effects on the privacy budget across different datasets, with reference values set at \{0.1, 0.2, 0.4, 0.8, 1\}. The privacy budget inversely correlates with $\lambda$, indicating stronger privacy as $\lambda$ decreases. Our experiments, conducted across ten different instantiations, report mean values that illustrate out method's resilience even with increasing noise, as shown in Fig.~\ref{fig:accuracy_dp}.

\begin{figure}[!t]
	\centering
	\includegraphics[width=0.65\linewidth]{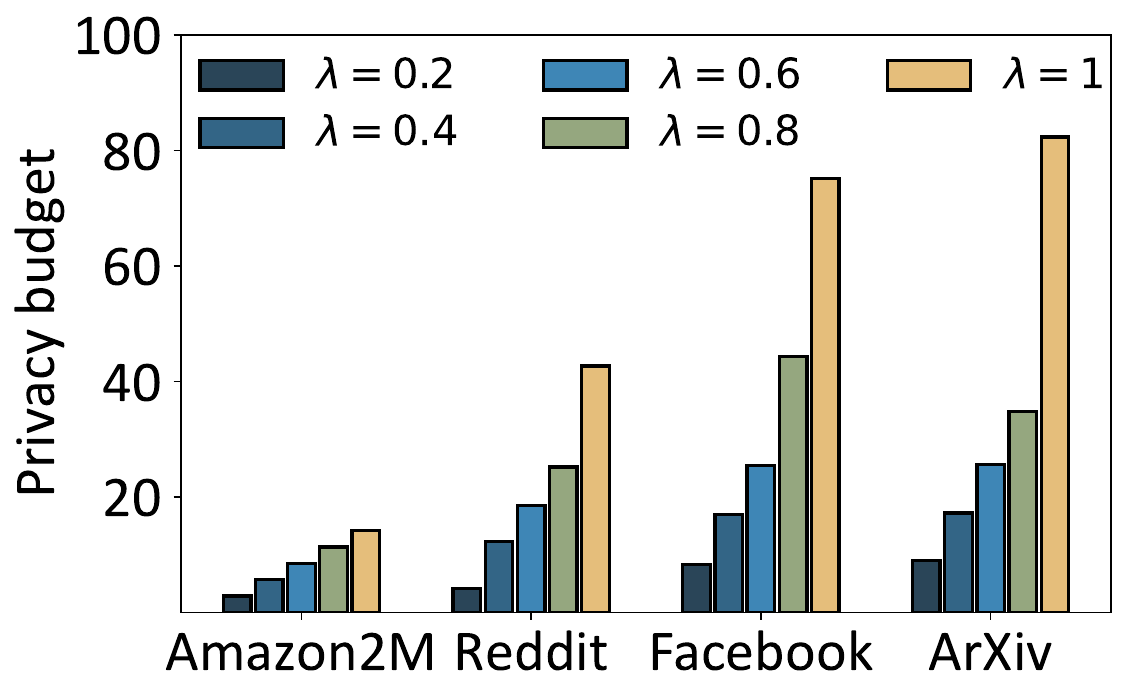}
	\caption{Privacy budget vs. noise ($\propto \frac{1}{\epsilon}$) for different datasets}
	\label{fig:budget_dp}
\end{figure}

The empirical results from Figs.~\ref{fig:accuracy_dp} and \ref{fig:budget_dp} offer a granular view of the performance of our model under varying levels of differential privacy noise. On the Amazon2M dataset, the accuracy only marginally decreases from 0.82 to 0.76 as $\lambda$ decreases from 1.0 to 0.2, despite the privacy budget increasing from 14.15 to 2.83. In contrast, the Facebook dataset exhibits a steeper decline in accuracy, from 0.85 to 0.37, for the same range of $\lambda$, suggesting that our method faces challenges in datasets with social network structures.
ArXiv's complex citation network structure demonstrates the capability of our approach to maintain a high level of accuracy with an accuracy decrease from 0.86 at $\lambda = 1.0$ to 0.74 at $\lambda = 0.2$, despite an increase in privacy budget from 82.29 to 82.98.  
This resilience in accuracy with a stringent $\delta$ of $10^{-5}$ is notable, highlighting the method's suitability for applications with complex data relationships.
These insights reveal the fine efficacy of our method in managing the trade-off between maintaining data utility and providing strong privacy guarantees, thereby validating its application in real-world scenarios where data sensitivity is paramount.

\begin{figure}[!h]
	\centering
	\includegraphics[width=0.7\linewidth]{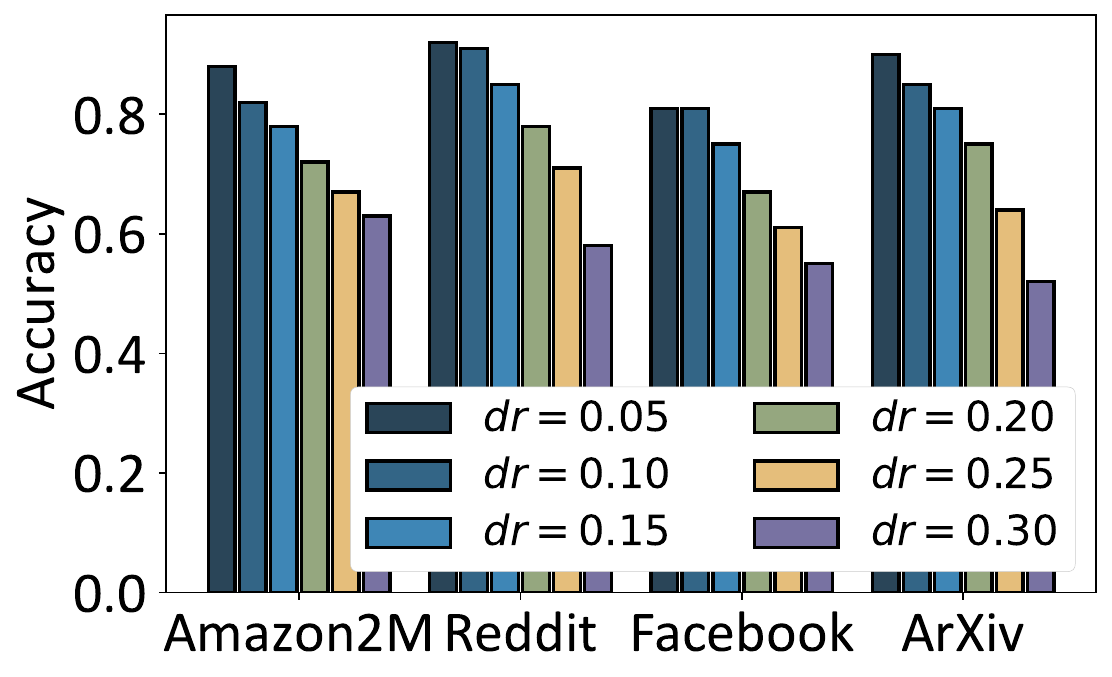}
	\caption{Accuracy under varying pruning dropping ratio (dr)}
	\label{fig:accuracy_dr}
	\vspace{-5mm}
\end{figure}

\subsection{Effectiveness of Graph Pruning}

\begin{figure*}[!t]
	\centering
	\scriptsize
	\subfloat[Accuracy]{\includegraphics[width=0.29\linewidth]{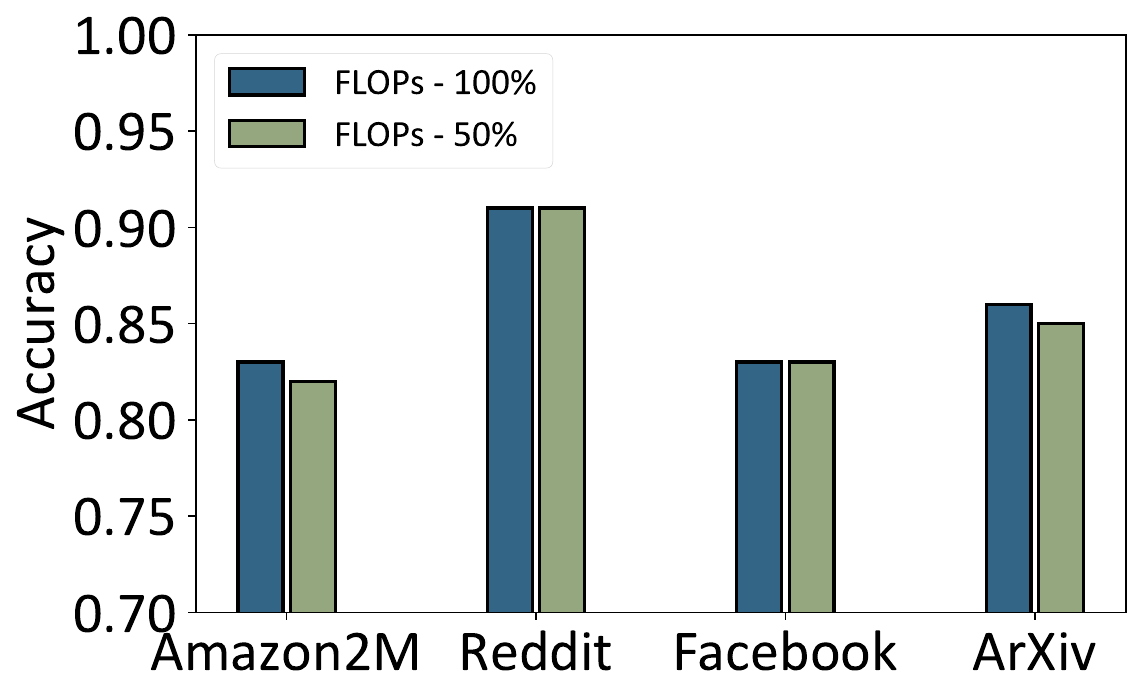}} \hspace{2mm}
	\subfloat[Latency]{\includegraphics[width=0.29\linewidth]{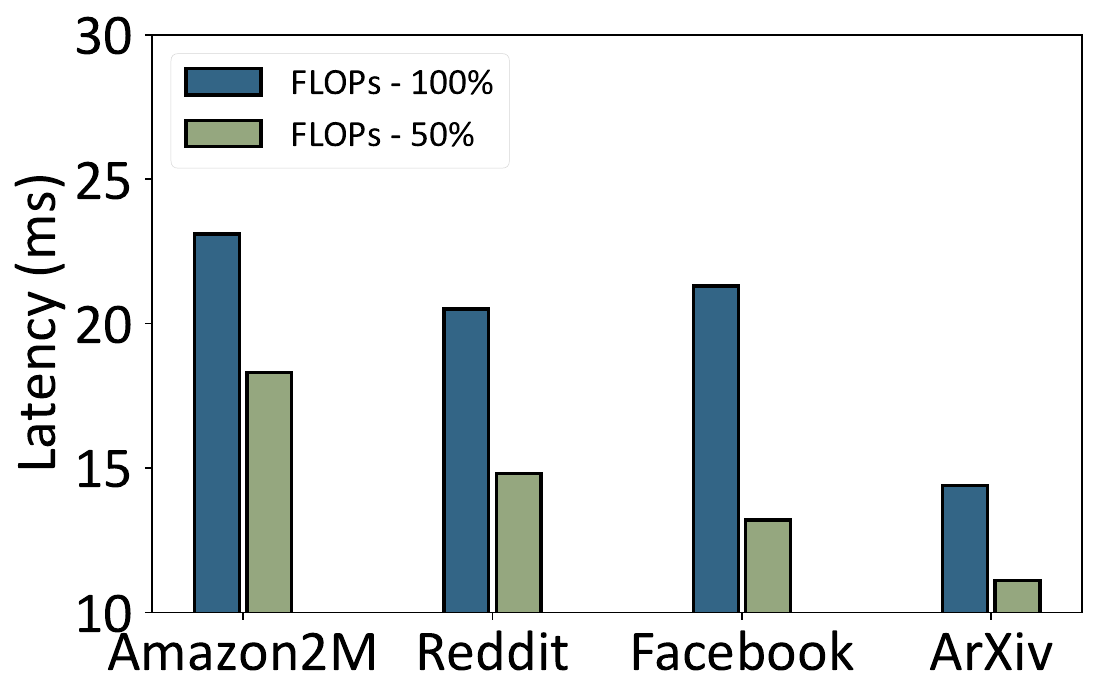}}\hspace{2mm}
	\subfloat[GPU Memory]{\includegraphics[width=0.30\linewidth]{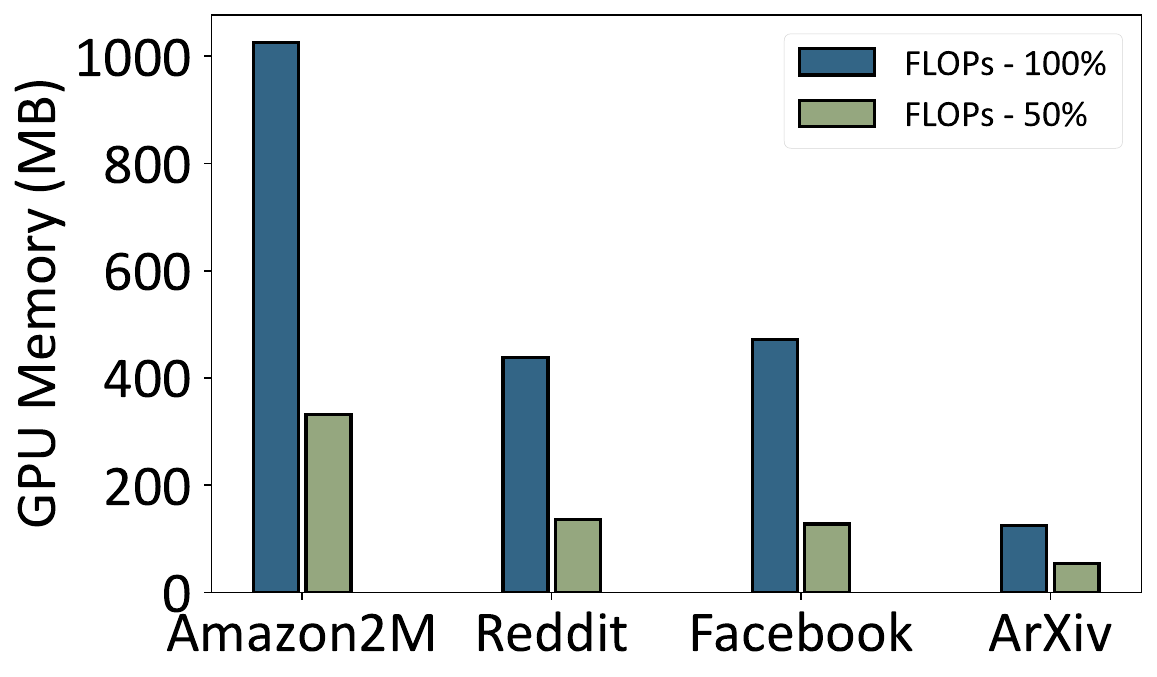}}
	\caption{Comparison of FLOPs-constrained model pruning under different datasets}
	\label{fig:eve_flops}
	\vspace{-6mm}
\end{figure*}
Our study also aims to assess the impact of different edge dropping ratios on the performance of GNNs.
Figure~\ref{fig:accuracy_dr} presents the accuracy of models across various datasets after applying pruning.

A key observation is the trend of diminishing accuracy with increased pruning, which is consistent across all datasets. For instance, on the Amazon2M dataset, the accuracy declines from 0.88 at a 5\% dropping ratio to 0.63 at 30\%. This trend indicates that while some level of pruning can be beneficial, excessive pruning might lead to the loss of critical information, negatively impacting the performance of the model.
The Reddit dataset shows high resilience to pruning, maintaining an accuracy above 0.9 even with a 10\% drop. In contrast, ArXiv experiences a more pronounced decrease, with accuracy falling to 0.52 at a 30\% dropping ratio. This highlights the varying sensitivities of different datasets to pruning, with some retaining essential information even in a sparser network, while others requiring a denser structure to maintain performance.
These findings suggest that optimal pruning ratios are dataset-dependent and should be chosen carefully to strike a balance between GNNs computational efficiency and accuracy retention.

\subsection{Effectiveness of Model Pruning}
\label{subsec:effectiveness_model_pruning}

Our analysis focuses on the trade-offs between computational efficiency and model performance post-pruning. As detailed in Figure~\ref{fig:eve_flops}, we fix the paramenter floating-point operations per second (FLOPs) at 50\% and report record computational efficiency across all datasets.

On Amazon2M, a 50\% reduction in FLOPs results in a slight decrease in accuracy from 0.83 to 0.82, but the inference latency is significantly reduced from 23.1ms to 18.3ms, and GPU memory usage drops sharply from 1025.7MB to 332.1MB. This indicates that substantial computational gains can be achieved with minimal impact on model accuracy.
For Reddit, the accuracy of the model remains stable at 0.91 despite a 50\% FLOPs reduction, showcasing the model resilience to pruning. Moreover, the latency and memory usage decrease significantly, highlighting the effectiveness of pruning in maintaining performance while enhancing efficiency.
Facebook's model shows a slight accuracy drop from 0.83 to 0.81 with halved FLOPs, suggesting a mild sensitivity to pruning. However, the reduction in latency and memory usage is noteworthy, indicating that the model benefits from computational efficiency improvements.
The ArXiv dataset displays remarkable stability in accuracy, decreasing only marginally from 0.86 to 0.85 with a 50\% reduction in FLOPs. The significant decrease in both latency and memory usage further underscores the benefits of pruning in complex datasets.

These results demonstrate that model pruning can significantly enhance computational efficiency across various datasets, with some variations in the degree to which accuracy is affected. This suggests a promising avenue for optimizing GNNs in computationally constrained environments.

\subsection{Complexity of Split Learning under Different Clients}

\label{subsec:comparative_analysis_fl_sl}

In our study we rigorously evaluated the complexity and efficiency of Federated Learning (FL) and SL in training GNNs across various client configurations and datasets. This analysis is meant to understand the scalability and applicability of these learning paradigms in distributed environments, particularly in limited-bandwidth satellite networks.

\begin{figure}[!h]
	\centering
	\includegraphics[width=0.65\linewidth]{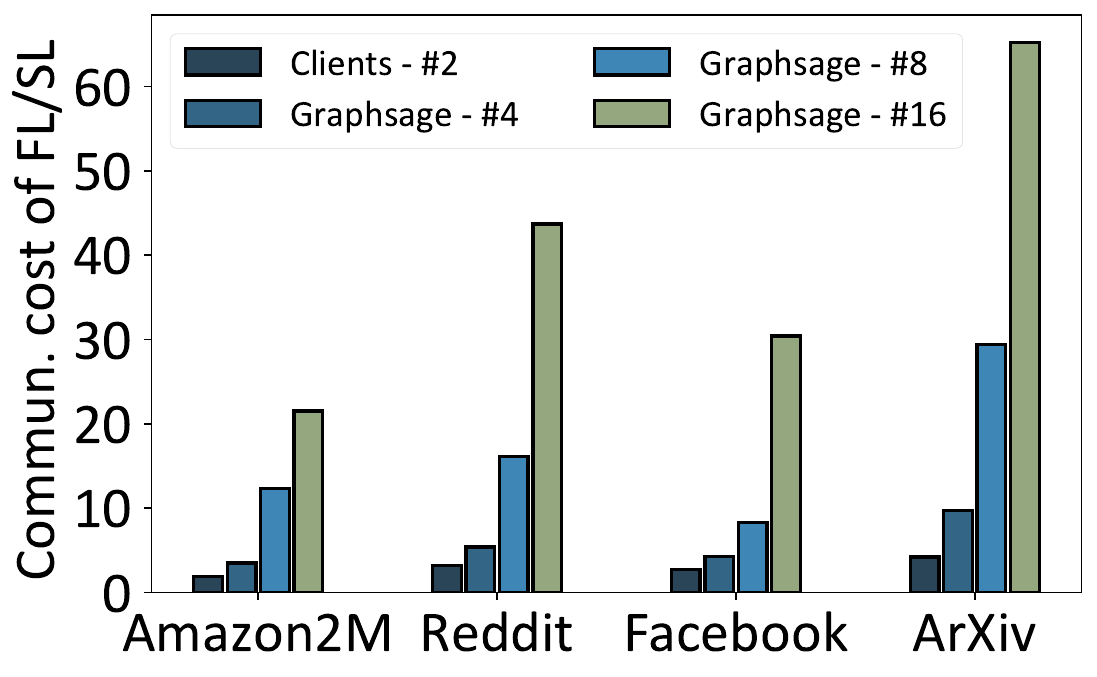}
	\caption{Comm. cost ratio of FL against split learning}
	\label{fig:flslcomp}
\end{figure}

\textbf{FL vs. SL Communication Costs.}
For FL, we measure communication costs by tracking data transmissions between clients and a central server during the model training process, as illustrated in Figure~\ref{fig:flslcomp}. The analysis reveals a significant increase in communication overhead with the addition of more clients. For example, in the Amazon2M dataset, communication costs rise from 1.9 units with 2 clients to 21.5 units with 16 clients. Similarly, Reddit's communication cost increases from 3.2 to 43.7 units, while Facebook and ArXiv experience substantial increases from 2.7 and 4.2 units to 30.4 and 65.2 units, respectively, under the same client scaling.

\textbf{Efficiency of SL-based GNNs.}
Our findings underscore the enhanced efficiency of SL-based GNNs over FL, particularly in scenarios involving a large number of clients. The communication cost ratios indicate that SL requirement for data transmission remains significantly lower than that of FL, despite the increasing number of clients. This efficiency gain is more pronounced in datasets like Facebook and ArXiv, where FL communication costs rise sharply with more clients.

\textbf{Implications for Resource-Constrained Environments.}
These results highlight the suitability of SL in bandwidth-limited environments like satellite networks. SL proves to be a more bandwidth-efficient approach compared to FL, especially as the complexity and number of clients increase.

In summary, the comparative analysis of FL and SL in GNNs reveals the advantages of SL in terms of communication efficiency, particularly in distributed learning scenarios with multiple clients and limited bandwidth resources.

\section{Conclusion}
Our work integrates GNNs into satellite-based distributed learning environments using SL and differential privacy. We propose a novel system design that addresses data privacy, computational efficiency, and network integrity challenges. Experimental results highlight the superior efficiency of SL over FL, particularly in scenarios with multiple clients or limited bandwidth. Additionally, our graph and model pruning techniques effectively refine GNNs for distributed learning. These contributions significantly advance distributed learning methodologies for the complex, resource-constrained settings of satellite communication networks. Future work will focus on enhancing the resilience and scalability of our system, developing models that adapt to network fluctuations, variable bandwidth conditions, and dynamic environments. Further experiments will explore a broader range of GNN architectures and diverse datasets, including unstructured and semi-structured data, to validate and extend our findings for satellite-based distributed learning systems.


\bibliographystyle{IEEEtran}
\bibliography{lib.bib}

\begin{thebibliography}{10}
\providecommand{\url}[1]{#1}
\csname url@samestyle\endcsname
\providecommand{\newblock}{\relax}
\providecommand{\bibinfo}[2]{#2}
\providecommand{\BIBentrySTDinterwordspacing}{\spaceskip=0pt\relax}
\providecommand{\BIBentryALTinterwordstretchfactor}{4}
\providecommand{\BIBentryALTinterwordspacing}{\spaceskip=\fontdimen2\font plus
\BIBentryALTinterwordstretchfactor\fontdimen3\font minus
  \fontdimen4\font\relax}
\providecommand{\BIBforeignlanguage}[2]{{%
\expandafter\ifx\csname l@#1\endcsname\relax
\typeout{** WARNING: IEEEtran.bst: No hyphenation pattern has been}%
\typeout{** loaded for the language `#1'. Using the pattern for}%
\typeout{** the default language instead.}%
\else
\language=\csname l@#1\endcsname
\fi
#2}}
\providecommand{\BIBdecl}{\relax}
\BIBdecl

\bibitem{yang20196g}
P.~Yang, Y.~Xiao, M.~Xiao, and S.~Li, ``6g wireless communications: Vision and
  potential techniques,'' \emph{IEEE Network}, vol.~33, no.~4, pp. 70--75,
  2019.

\bibitem{qian2021multi}
L.~Qian, P.~Yang, Y.~L. Guan, Z.~Liu, Y.~Xiao, K.~Jiang, and M.~Xiao,
  ``Multi-dimensional polarized modulation for land mobile satellite
  communications,'' \emph{IEEE Transactions on Cognitive Communications and
  Networking}, 2021.

\bibitem{zhu2018dual}
J.~Zhu, P.~Yang, Y.~Xiao, M.~Di~Renzo, and S.~Li, ``Dual polarized spatial
  modulation for land mobile satellite communications,'' in \emph{IEEE Globecom
  Workshops}.\hskip 1em plus 0.5em minus 0.4em\relax IEEE, 2018.

\bibitem{Koumaras2019eucnc}
H.~Koumaras, T.~Anagnostopoulos, M.~Kourtis, G.~Gardikis, N.~Papadakis,
  A.~Perentos, M.~Fotiou, A.~Phinikarides, M.~Georgiades, V.~Frascolla, and
  D.~Tsolkas, ``5g experimentation facility supporting satellite-terrestrial
  integration: The 5genesis approach,'' \emph{European Conference on Networks
  and Communications (EuCNC)}, 2019.

\bibitem{qu2024mobile}
G.~Qu, Q.~Chen, W.~Wei, Z.~Lin, X.~Chen, and K.~Huang, ``Mobile edge
  intelligence for large language models: A contemporary survey,'' \emph{arXiv
  preprint arXiv:2407.18921}, 2024.

\bibitem{wang2024graph}
X.~Wang, K.~Guan, D.~He, A.~Hrovat, R.~Liu, Z.~Zhong, A.~Al-Dulaimi, and K.~Yu,
  ``Graph neural network enabled propagation graph method for channel
  modeling,'' \emph{IEEE Transactions on Vehicular Technology}, 2024.

\bibitem{moorthy2024survey}
S.~K. Moorthy and J.~Jagannath, ``Survey of graph neural network for internet
  of things and nextg networks,'' \emph{arXiv preprint arXiv:2405.17309}, 2024.

\bibitem{verbraeken2020survey}
J.~Verbraeken, M.~Wolting, J.~Katzy, J.~Kloppenburg, T.~Verbelen, and J.~S.
  Rellermeyer, ``A survey on distributed machine learning,'' \emph{Acm
  Computing Surveys}, 2020.

\bibitem{DBLPSong_Lyu}
S.~Lyu, Z.~Lin, G.~Qu, X.~Chen, X.~Huang, and P.~Li, ``Optimal resource
  allocation for u-shaped parallel split learning,'' \emph{arXiv preprint
  arXiv:2308.08896}, 2023.

\bibitem{lin2024adaptsfl}
Z.~Lin, G.~Qu, W.~Wei, X.~Chen, and K.~K. Leung, ``Adaptsfl: Adaptive split
  federated learning in resource-constrained edge networks,'' \emph{arXiv
  preprint arXiv:2403.13101}, 2024.

\bibitem{qu2024trimcaching}
G.~Qu, Z.~Lin, F.~Liu, X.~Chen, and K.~Huang, ``Trimcaching: Parameter-sharing
  ai model caching in wireless edge networks,'' \emph{arXiv preprint
  arXiv:2405.03990}, 2024.

\bibitem{lin2024splitlora}
Z.~Lin, X.~Hu, Y.~Zhang, Z.~Chen, Z.~Fang, X.~Chen, A.~Li, P.~Vepakomma, and
  Y.~Gao, ``Splitlora: A split parameter-efficient fine-tuning framework for
  large language models,'' \emph{arXiv preprint arXiv:2407.00952}, 2024.

\bibitem{mammen2021federated}
P.~M. Mammen, ``Federated learning: Opportunities and challenges,'' \emph{arXiv
  preprint arXiv:2101.05428}, 2021.

\bibitem{kodheli2020satellite}
O.~Kodheli, E.~Lagunas, N.~Maturo, S.~K. Sharma, B.~Shankar, J.~F.~M. Montoya,
  J.~C.~M. Duncan, D.~Spano, S.~Chatzinotas, S.~Kisseleff \emph{et~al.},
  ``Satellite communications in the new space era: A survey and future
  challenges,'' \emph{IEEE Communications Surveys \& Tutorials}, 2020.

\bibitem{fang2024automated}
Z.~Fang, Z.~Lin, Z.~Chen, X.~Chen, Y.~Gao, and Y.~Fang, ``Automated federated
  pipeline for parameter-efficient fine-tuning of large language models,''
  \emph{arXiv preprint arXiv:2404.06448}, 2024.

\bibitem{lin2023pushing}
Z.~Lin, G.~Qu, Q.~Chen, X.~Chen, Z.~Chen, and K.~Huang, ``Pushing large
  language models to the 6g edge: Vision, challenges, and opportunities,''
  \emph{arXiv preprint arXiv:2309.16739}, 2023.

\bibitem{lin2024split}
Z.~Lin, G.~Qu, X.~Chen, and K.~Huang, ``Split learning in 6g edge networks,''
  \emph{IEEE Wireless Communications}, 2024.

\bibitem{lin2023fedsn}
Z.~Lin, Z.~Chen, Z.~Fang, X.~Chen, X.~Wang, and Y.~Gao, ``Fedsn: A general
  federated learning framework over leo satellite networks,'' \emph{arXiv
  preprint arXiv:2311.01483}, 2023.

\bibitem{lin2024efficient}
Z.~Lin, G.~Zhu, Y.~Deng, X.~Chen, Y.~Gao, K.~Huang, and Y.~Fang, ``Efficient
  parallel split learning over resource-constrained wireless edge networks,''
  \emph{IEEE Transactions on Mobile Computing}, 2024.

\bibitem{wang2021grouting}
H.~Wang, Y.~Ran, L.~Zhao, J.~Wang, J.~Luo, and T.~Zhang, ``Grouting: dynamic
  routing for leo satellite networks with graph-based deep reinforcement
  learning,'' in \emph{International Conference on Hot Information-Centric
  Networking}, 2021.

\bibitem{chiang2019cluster}
W.-L. Chiang, X.~Liu, S.~Si, Y.~Li, S.~Bengio, and C.-J. Hsieh, ``Cluster-gcn:
  An efficient algorithm for training deep and large graph convolutional
  networks,'' in \emph{ACM SIGKDD international conference on knowledge
  discovery \& data mining}, 2019.

\bibitem{hu2020open}
W.~Hu, M.~Fey, M.~Zitnik, Y.~Dong, H.~Ren, B.~Liu, M.~Catasta, and J.~Leskovec,
  ``Open graph benchmark: Datasets for machine learning on graphs,''
  \emph{Advances in Neural Information Processing Systems}, 2020.

\bibitem{hamilton2017inductive}
W.~Hamilton, Z.~Ying, and J.~Leskovec, ``Inductive representation learning on
  large graphs,'' \emph{Advances in Neural Information Processing Systems},
  vol.~30, 2017.

\bibitem{poirot2019split}
M.~G. Poirot, P.~Vepakomma, K.~Chang, J.~Kalpathy-Cramer, R.~Gupta, and
  R.~Raskar, ``Split learning for collaborative deep learning in healthcare,''
  \emph{arXiv preprint arXiv:1912.12115}, 2019.

\bibitem{vepakomma2018split}
P.~Vepakomma, O.~Gupta, T.~Swedish, and R.~Raskar, ``Split learning for health:
  Distributed deep learning without sharing raw patient data,'' \emph{arXiv
  preprint arXiv:1812.00564}, 2018.

\bibitem{gao2020end}
Y.~Gao, M.~Kim, S.~Abuadbba, Y.~Kim, C.~Thapa, K.~Kim, S.~A. Camtepe, H.~Kim,
  and S.~Nepal, ``End-to-end evaluation of federated learning and split
  learning for internet of things,'' \emph{arXiv preprint arXiv:2003.13376},
  2020.

\bibitem{koda2019one}
Y.~Koda, J.~Park, M.~Bennis, K.~Yamamoto, T.~Nishio, and M.~Morikura, ``One
  pixel image and rf signal based split learning for mmwave received power
  prediction,'' in \emph{International Conference on Emerging Networking
  EXperiments and Technologies}, 2019.

\bibitem{mcmahan2017communication}
B.~McMahan, E.~Moore, D.~Ramage, S.~Hampson, and B.~A. y~Arcas,
  ``Communication-efficient learning of deep networks from decentralized
  data,'' in \emph{Artificial intelligence and statistics}, 2017.

\bibitem{bonawitz2019towards}
K.~Bonawitz, H.~Eichner, W.~Grieskamp, D.~Huba, A.~Ingerman, V.~Ivanov,
  C.~Kiddon, J.~Kone{\v{c}}n{\`y}, S.~Mazzocchi, B.~McMahan \emph{et~al.},
  ``Towards federated learning at scale: System design,'' \emph{Proceedings of
  Machine Learning and Systems}, 2019.

\bibitem{konevcny2016federated}
J.~Kone{\v{c}}n{\`y}, H.~B. McMahan, D.~Ramage, and P.~Richt{\'a}rik,
  ``Federated optimization: Distributed machine learning for on-device
  intelligence,'' \emph{arXiv preprint arXiv:1610.02527}, 2016.

\bibitem{gupta2018distributed}
O.~Gupta and R.~Raskar, ``Distributed learning of deep neural network over
  multiple agents,'' \emph{Journal of Network and Computer Applications}, 2018.

\bibitem{sattler2019robust}
F.~Sattler, S.~Wiedemann, K.-R. M{\"u}ller, and W.~Samek, ``Robust and
  communication-efficient federated learning from non-iid data,'' \emph{IEEE
  Transactions on Neural Networks and Learning Systems}, 2019.

\bibitem{wang2020optimize}
C.~Wang, X.~Wei, and P.~Zhou, ``Optimize scheduling of federated learning on
  battery-powered mobile devices,'' in \emph{IEEE International Parallel and
  Distributed Processing Symposium (IPDPS)}, 2020.

\bibitem{zhang2022advancing}
Y.~Zhang, Y.~Yao, P.~Ram, P.~Zhao, T.~Chen, M.~Hong, Y.~Wang, and S.~Liu,
  ``Advancing model pruning via bi-level optimization,'' \emph{Advances in
  Neural Information Processing Systems}, 2022.

\bibitem{han2015learning}
S.~Han, J.~Pool, J.~Tran, and W.~Dally, ``Learning both weights and connections
  for efficient neural network,'' \emph{Advances in Neural Information
  Processing Systems}, 2015.

\bibitem{li2016pruning}
H.~Li, A.~Kadav, I.~Durdanovic, H.~Samet, and H.~P. Graf, ``Pruning filters for
  efficient convnets,'' \emph{arXiv preprint arXiv:1608.08710}, 2016.

\bibitem{liu2017learning}
Z.~Liu, J.~Li, Z.~Shen, G.~Huang, S.~Yan, and C.~Zhang, ``Learning efficient
  convolutional networks through network slimming,'' in \emph{IEEE/CVF
  International Conference on Computer Vision}, 2017.

\bibitem{he2019filter}
Y.~He, P.~Liu, Z.~Wang, Z.~Hu, and Y.~Yang, ``Filter pruning via geometric
  median for deep convolutional neural networks acceleration,'' in
  \emph{IEEE/CVF International Conference on Computer Vision}, 2019.

\bibitem{lin2020hrank}
M.~Lin, R.~Ji, Y.~Wang, Y.~Zhang, B.~Zhang, Y.~Tian, and L.~Shao, ``Hrank:
  Filter pruning using high-rank feature map,'' in \emph{IEEE/CVF International
  Conference on Computer Vision}, 2020.

\bibitem{he2018amc}
Y.~He, J.~Lin, Z.~Liu, H.~Wang, L.-J. Li, and S.~Han, ``Amc: Automl for model
  compression and acceleration on mobile devices,'' in \emph{European
  conference on computer vision (ECCV)}, 2018.

\bibitem{liu2019metapruning}
Z.~Liu, H.~Mu, X.~Zhang, Z.~Guo, X.~Yang, K.-T. Cheng, and J.~Sun,
  ``Metapruning: Meta learning for automatic neural network channel pruning,''
  in \emph{IEEE/CVF International Conference on Computer Vision}, 2019.

\bibitem{li2020eagleeye}
B.~Li, B.~Wu, J.~Su, and G.~Wang, ``Eagleeye: Fast sub-net evaluation for
  efficient neural network pruning,'' in \emph{European Conference on Computer
  Vision}, 2020.

\bibitem{lin2020channel}
M.~Lin, R.~Ji, Y.~Zhang, B.~Zhang, Y.~Wu, and Y.~Tian, ``Channel pruning via
  automatic structure search,'' \emph{arXiv preprint arXiv:2001.08565}, 2020.

\bibitem{kipf2016semi}
T.~N. Kipf and M.~Welling, ``Semi-supervised classification with graph
  convolutional networks,'' \emph{arXiv preprint arXiv:1609.02907}, 2016.

\bibitem{velivckovic2017graph}
P.~Veli{\v{c}}kovi{\'c}, G.~Cucurull, A.~Casanova, A.~Romero, P.~Lio, and
  Y.~Bengio, ``Graph attention networks,'' \emph{arXiv preprint
  arXiv:1710.10903}, 2017.

\bibitem{qu2022blockchain}
Y.~Qu, M.~P. Uddin, C.~Gan, Y.~Xiang, L.~Gao, and J.~Yearwood,
  ``Blockchain-enabled federated learning: A survey,'' \emph{ACM Computing
  Surveys}, vol.~55, no.~4, pp. 1--35, 2022.

\bibitem{xu2023asynchronous}
C.~Xu, Y.~Qu, Y.~Xiang, and L.~Gao, ``Asynchronous federated learning on
  heterogeneous devices: A survey,'' \emph{Computer Science Review}, vol.~50,
  p. 100595, 2023.

\bibitem{dwork2006differential}
C.~Dwork, ``Differential privacy,'' in \emph{International colloquium on
  automata, languages, and programming}, 2006.

\bibitem{traud2012social}
A.~L. Traud, P.~J. Mucha, and M.~A. Porter, ``Social structure of facebook
  networks,'' \emph{Physica A: Statistical Mechanics and its Applications},
  2012.

\bibitem{kingma2014adam}
D.~P. Kingma and J.~Ba, ``Adam: A method for stochastic optimization,''
  \emph{arXiv preprint arXiv:1412.6980}, 2014.

\end{thebibliography}

\vfill

\end{document}